\documentclass{svproc}
\usepackage{url}
\usepackage{graphicx}

\begin{document}
\mainmatter              
\title{A Robotic System to Automate the Disassembly of PCB Components}
\titlerunning{A Robotic System to Automate the Disassembly of PCB Components} 
%
\author{Silvia Santos\inst{1} \and Lino Marques\inst{1}
	\and
	Pedro Neto\inst{2}}
\authorrunning{Silvia Santos et al.} 
%
\tocauthor{Silvia Santos, Lino Marques, and Pedro Neto}
\institute{Institute of Systems and Robotics, Department of Electrical and Computer Engineering, University of Coimbra, 3030-290 Coimbra, Portugal,\\
	\and
	University of Coimbra, CEMMPRE, ARISE, Department of Mechanical Engineering, 3030-788, Coimbra, Portugal \\ \email{pedro.neto@dem.uc.pt}\\ } 

\maketitle              

\begin{abstract}
The disposal and recycling of electronic waste (e-waste) poses a significant global challenge. The disassembly of components is a crucial step in achieving an efficient recycling process, avoiding destructive methods. While manual disassembly remains the norm due to the diversity and complexity of components, there is a growing interest in automating the process to enhance efficiency and reduce labor costs. This study endeavors to automate the desoldering process and the extraction of components from printed circuit boards (PCBs) by implementing a robotic solution. The proposed strategy encompasses multiple phases, one of which involves the precise contact of the developed robotic tool with the components on the PCB. The tool was designed to exert a controlled force on the PCB component, thereby efficiently desoldering it from the board. Results demonstrate the feasibility of achieving a high success rate in removing PCB components, including the ones from mobile phones. The desoldering  success rate observed is approximately 100\% for the larger components.

\end{abstract}

\section{Introduction}
The treatment of electronic waste (e-waste) is a global challenge \cite{Adeniyi20221}. By the year 2030, the projected amount of waste generated from electrical and electronic equipment is anticipated to surpass 75 million tons \cite{Hongguo20226}. Given this scenario, there is a paramount need to create economically viable technologies aimed at recycling and recovering end-of-life electrical and electronic equipment.

\subsection{PCB Dismantling }

Dismantling or disassembly is widely regarded by numerous researchers as the initial and pivotal stage among all the phases of product recovery. This is due to the fact that the effectiveness of this operation has the potential to influence the outcome of subsequent stages in the recycling process \cite{Bernd20162}. An effective dismantling process can also help mitigate the excessive utilization of reagents in chemical procedures, which are employed to isolate various valuable metals. Additionally, it presents an alternative approach to destructive techniques like crushing, which, although quicker, restricts the usability of only specific recycled materials due to its detrimental impact on the structural integrity of all components \cite{Ange20218}.

Recent research indicates that most disassembly tasks continue to rely on manual labor. This can be attributed to the wide-ranging and distinct characteristics of products, requiring distinct handling methods and extraction tools. Among the promising approaches for automating this process is robotic disassembly, which offers a notable advantage stemming from its capacity to execute repetitive tasks. This capability contributes to reduced operation durations and overall costs associated with the process \cite{Li20197}. Nonetheless, the full implementation of robotic disassembly requires further advancements, given the existing challenges related to robot autonomy, perception, and manipulation tools.

Mobile phones are quintessential electronic devices with ever-decreasing lifecycles, quickly seen as obsolete by consumers. Furthermore, due to their popularity and constant presence into users' daily routines, they stand as a significant contributor to the prevailing electronic waste dilemma \cite{Hongguo20226}. Another pertinent facet is that mobile phones exhibit distinct characteristics that introduce heightened complexities to the automated dismantling process compared to other devices. These include their compact size, the close proximity of components, and the non-uniform arrangement of elements on the PCB. Notably, the separation of PCB components plays a pivotal role in streamlining the recycling procedure. Furthermore, as PCBs serve as the foundation of electric and electronic devices, the successful dismantling of mobile phone PCBs holds the potential to extend their application across a broader spectrum of devices \cite{Chao201710}.

\subsection{Proposed Approach}
\begin{figure}
	\centering
	\includegraphics[scale = 0.5]{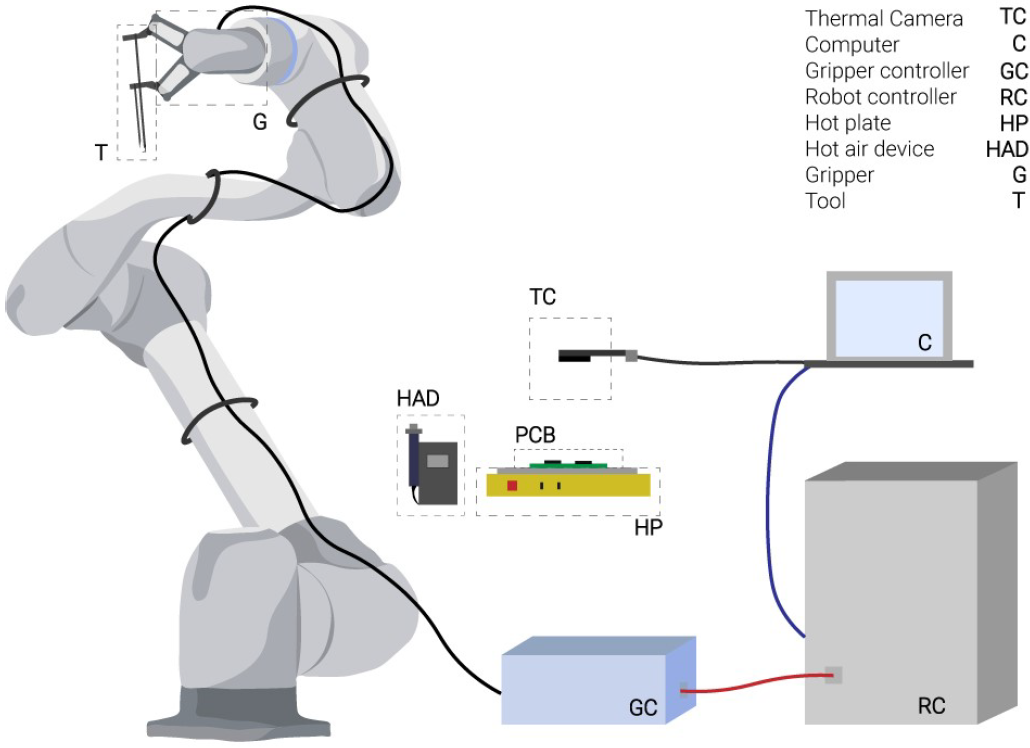}
	\caption{Schematic of the system highlighting the main hardware elements.}
	\label{fig1}
\end{figure}

This paper introduces a groundbreaking robotic system specifically designed for the desoldering process of PCB components, Fig. \ref{fig1}. The primary objective is to disengage PCB components from the board while preserving their structural integrity. A novel robotic tool is proposed to accomplish this task through a precisely applied contact force by the robot. Various manipulation strategies were thoroughly examined and evaluated. Through a series of experimental tests, the efficacy of the proposed solution was demonstrated.

\section{Background and Related Studies}

The process of extracting components from PCBs necessitates the robot tool’s contact with both the board and its individual components. Consequently, it becomes imperative to control the forces and positions of the robot to ensure the safety and efficacy of the operation. The tools employed for desoldering, grasping, and transporting PCB components face a series of obstacles due to the relatively small size of PCB components. The automation of this procedure requires the integration of diverse components into architectures that ensure their functionality, namely real-time force and motion control mechanisms for the robot.

Passive and active force control represent two primary methodologies for force control in robotics. Passive control involves controlling forces to stay within a predefined range that ensures the accomplishment of a specific task. Active control adapts contact forces based on the level of efficiency needed for a particular task, requiring knowledge of their magnitude. Thus, knowing in real-time the contact forces is imperative to achieve effective force control \cite{Mendes2013}. Active control can be categorized into direct and indirect approaches. In direct control, force and position control are performed independently \cite{Calanca2016A111}. Indirect control aims to achieve a desired compliant dynamic behavior of the robot's end-effector in the presence of interaction with the environment \cite{Calanca2016A111}. Impedance and admittance control emerge as two forms of indirect control \cite{Sadun2016A113}. Explicit implementations employ force sensors to measure the contact force values from the environment, while implicit implementations calculate these forces via other derived quantities. To implement robotic processes involving the contact of the robotic system with the surrounding environment, such as polishing or PCB disassembly, the robot compliance allows the robot to adapt to the surrounding space \cite{Huang2019A112,Su2022Design}. The grasping and manipulation of the PCB components can be accomplished using different grippers or customized tools. They can be mechanically actuated, featuring multiple contact points, or relying on methods like suction or magnetic actuation \cite{Gualtiero2014A22,RUGGERI2017441A222}. 

In the context of automated dismantling of e-waste, particularly the extraction of PCBs from mobile phones, the European project ADIR has proposed a method encompassing various phases for the disassembly process, namely image processing, robotic manipulation and the automated sorting of components into distinct fractions \cite{AIRD2017A28}. Recently, researchers have developed an automated disassembly system to recover central processing units (CPUs) from cell phones, \cite{He_2020}. This system employs eye-to-hand visual servoing, hot air for desoldering and a gripper for transportation of CPUs. The system's performance was evaluated through experimental tests using cell phones from various brands. In 2019, Apple introduced a machine designed to automatically dismantle cell phones. This machine segregates components based on their constituent materials \cite{Jie2022A31}. However, a challenge remains, the lack of a method to separate components from circuit boards without causing major damage to their physical integrity.

\section{Robotic desoldering process}
The robot is a collaborative robot with six degrees of freedom and integrated torque sensors in all joints (M1013, Doosan Robotics). Force sensing values at the end-effector are calculated based on the robot’s rigid-body dynamics and measured torque values. A set of software libraries and tools for building robot applications based on Robot Operating System (ROS) were used to develop and implement the running algorithms. The robot’s path planning was based on the kinematics considerations implemented into the robot controller. To heat the PCBs, a hot air device and a heated plate were employed. The robot is equipped with a two fingers gripper (RG6, OnRobot) and a customized tool to desolder and transport the PCB components, Fig. \ref{fig1}. The process to isolate the PCB components from the board, grasp them, and moving the components to a target location can be described by the following phases:

\begin{enumerate}
	\item Approach: The robot tool approaches the PCB component;
	\item Contact: The tool contacts with the component and it is controlled to reach and maintain a desired set force; 
	\item Melting: The melting point of the solder material is reached;
	\item Grasping: The robotic system grasps and holds the component;
	\item Transport: The component is moved to a target pose; 
	\item Release: The component is released.
\end{enumerate}

\subsection{Robot control}

\begin{figure}
	\centering
	\includegraphics[scale = 0.53]{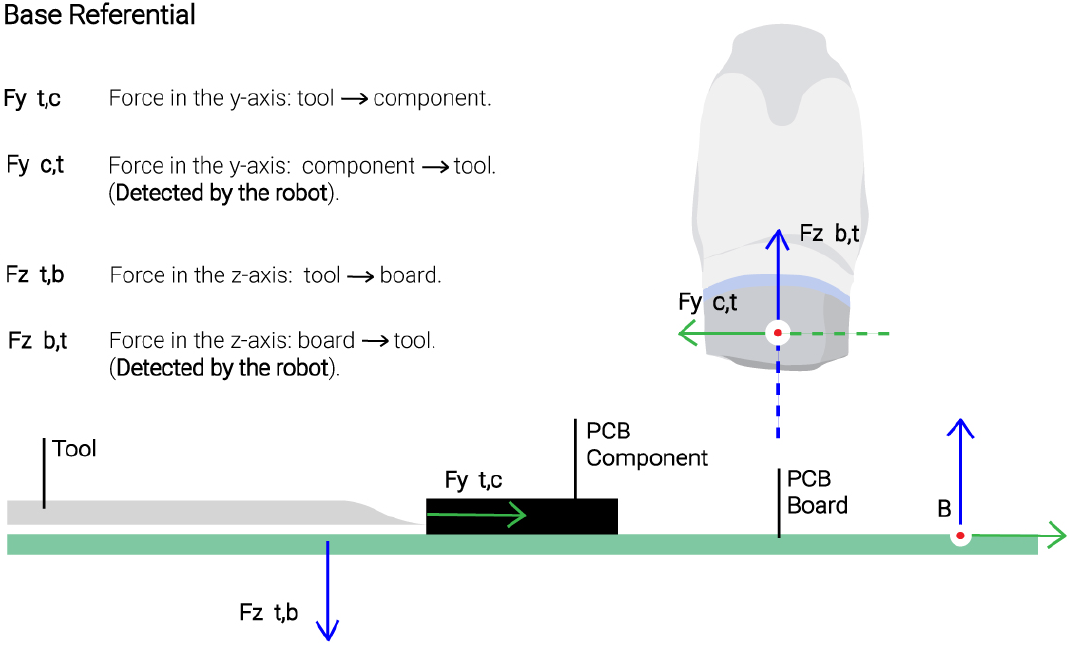}
	\caption{Representation of the forces acting on the tool during the contact with the PCB component and its representation at the robot end-effector level.}
	\label{fig2}
\end{figure}

\begin{figure}
	\centering
	\includegraphics[scale = 0.37]{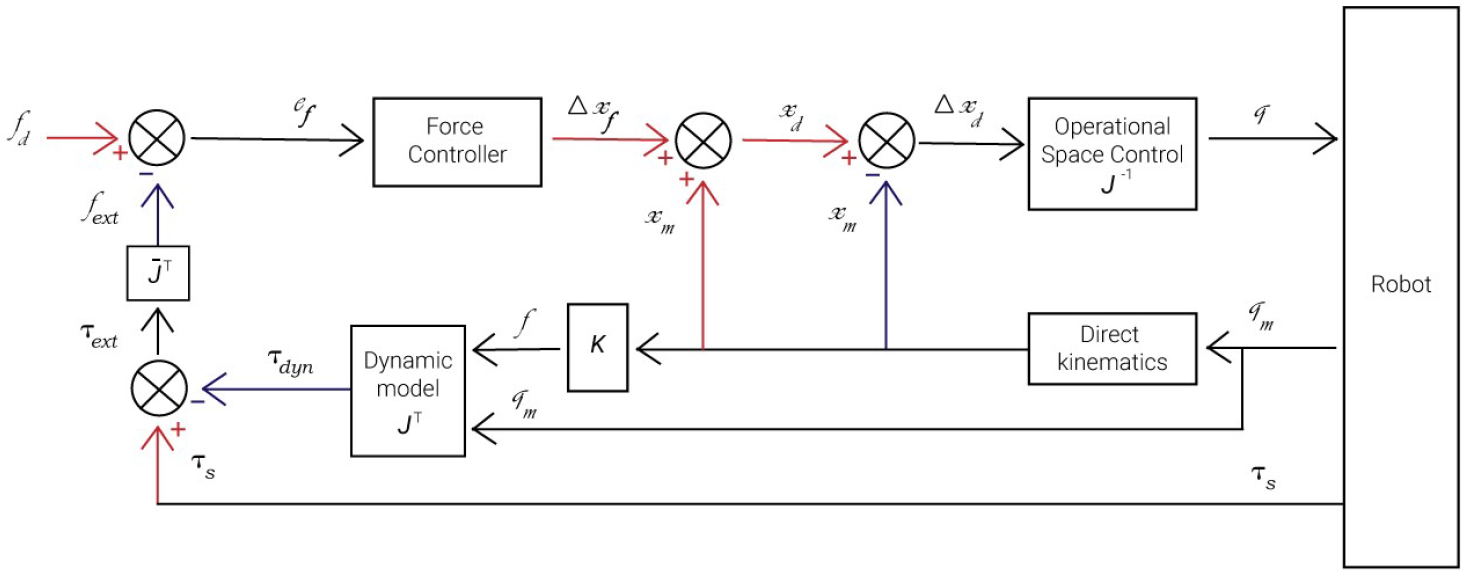}
	\caption{Robot control architecture.}
	\label{fig2-5}
\end{figure}

The robot force control was defined at the Cartesian level, setting a desired value, $f_d$, for the robot’s force in a specific direction according to the extraction direction of the component of interest. A compliance control was defined in the other directions because it is expected that the tool, at some point, touches other smaller components on the PCB. As such, it is helpful that the robot can make small changes in the path (according to these directions) to adapt to the main direction of motion to remove the component. Fig. \ref{fig2} shows a schematic of the forces applied to a given component in a PCB. The initial contact between the tool and the plate is detected by the force along the z-axis, ${F_{z b,t}}$. The solder melting and, consequently, the movement of the component is detected by the decrease of the y-axis force, ${F_{y c,t}}$. The proposed control architecture is detailed in Fig. \ref{fig2-5}.

\subsection{Tool}

\begin{figure}
	\centering
	\includegraphics[scale = 0.60]{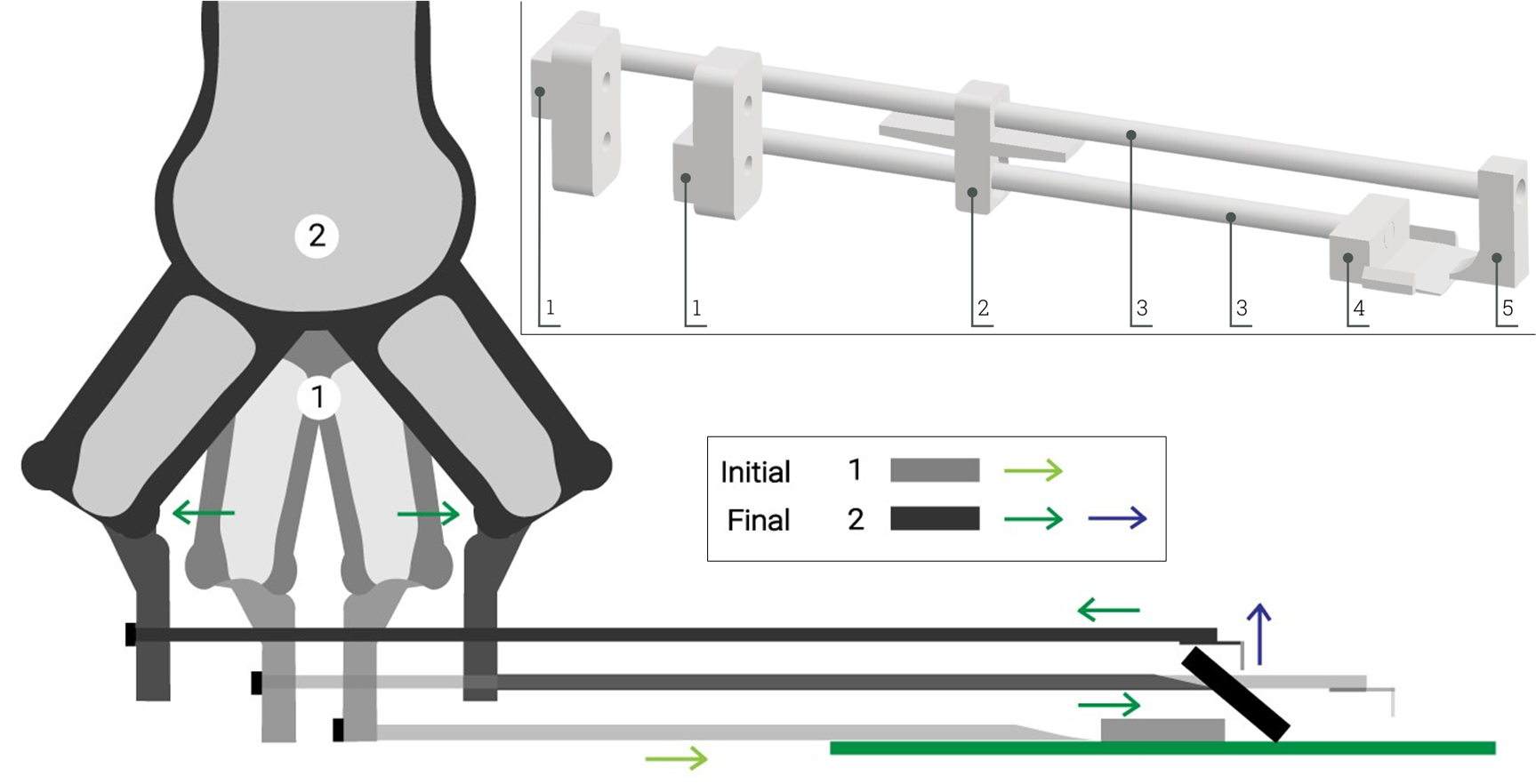}
	\caption{Schematic of the push-pull mechanism in which the tool connectors (1) are attached to the gripper fingers. Through contact, the tool can desolder the PCB component and grasp it. The tool main design elements are the connectors to the gripper fingers (1), the sliding mechanism that allows the lower and upper bars to slide at a set distance, the lower and upper bars (3), the element that contacts with the PCB component to desolder it (4), and the element that ensures the grasping (5).}
	\label{fig3}
\end{figure}

The main challenge related to the tool design was the ability of the tool to contact with the PCB components to desolder and grasp them. These tasks are increasingly challenging as the PCB components are smaller. It was defined that the tool will be attached to a two fingers gripper attached to the robot flange, being actuated by the open/close motion of the fingers to grasp the components, Fig. \ref{fig3}. When connected to the gripper fingers, the tool mechanism is identical to a push-pull system.

\section{Results and discussion}
\begin{figure}
	\centering
	\includegraphics[scale = 0.63]{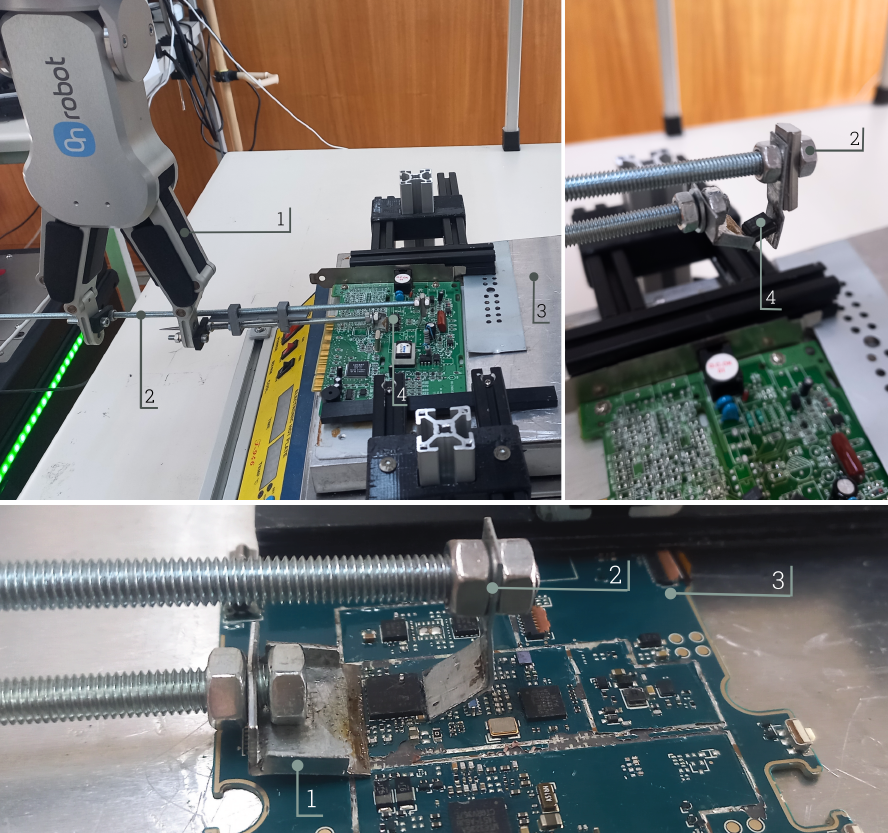}
	\caption{Experimental setup of the system highlighting the in-contact desoldering of a PCB component (left) and the grasping of such component (right). The system is composed by the gripper (1), the tool (2), the hot plate (3) and the PCB component (4). The bottom figure shows the PCB of a mobile phone.}
	\label{fig4}
\end{figure}

The proposed robotic system was tested and evaluated in the process of desoldering PCB components of multiple sizes, Fig. \ref{fig4}, discussing the performance of the system’s tool and the force control strategy. The contact area of the tool with the PCB component demonstrated effective but needs to be adjusted according to the size of the PCB component, otherwise, it may collide with other components. As can be seen in Fig. \ref{fig4}, some components are very close to each other. In such a context, the tool element that contacts with the PCB component and the one that ensures the grasping, Fig. \ref{fig3}, need to be customized according to the size of the components to desolder. Overall, the proposed tool design is capable to establish an appropriate contact with the PCB components, tolerating the temperatures produced by the hot air application. Nevertheless, it frequently shows instability during the grasping of the PCB component. 

\begin{figure}
	\centering
	\includegraphics[scale = 0.7]{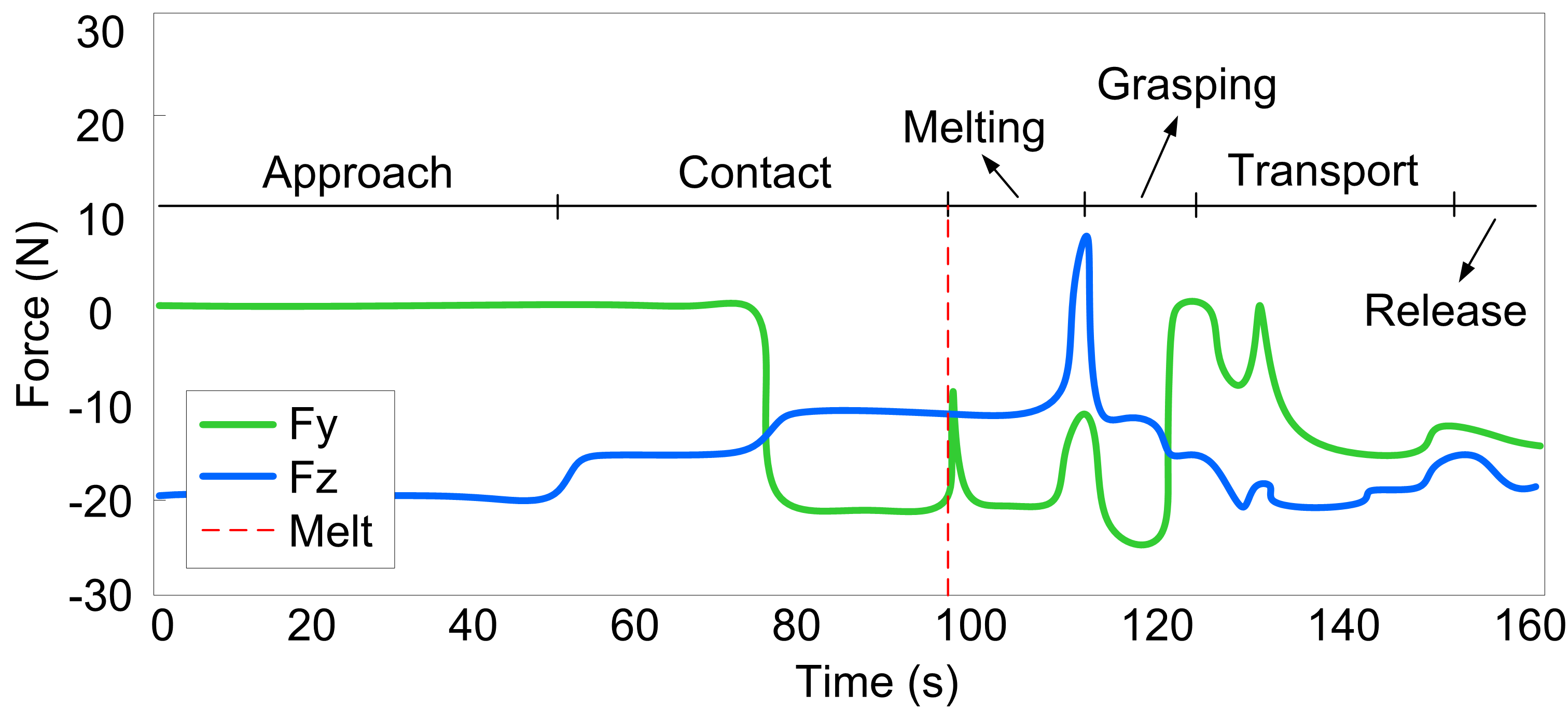}
	\caption{Measured force values along all the phases of the process. In this specific case, the extraction of a chip from a PCB board.}
	\label{fig5}
\end{figure}

The graph in Fig. \ref{fig5} shows the forces sensed by the tool during the different phases of the process, from the approach to the release of the PCB component. Only the forces along the y and z axis, $f_y$ and $f_z$, are represented as they are the most significant. Initially, during the approach phase, the sensed force along z axis is due to the weight of the gripper and tool. At the contact phase, the system detects a change in vertical force, due to the contact of the tool with the PCB, and then the force along y axis stabilizes at -20 N, which is the desired force we defined for the contact between the tool and the component (this is an indication that contact has been made between the tool and the board). This is an indication that the force control is effective. After a period, we reach the initial point of the melting phase, where $f_y$ suddenly decreases as an indication of the solder melting, and immediately the force controller quickly acts to keep the set force, finishing the melting process and consequently the removal of the component from the board. The component is completely desoldered after 10 seconds, depending on the amount of solder that connected the component to the board and the presence of components in the component’s seams preventing it from moving forward. Finally, the force controller is disabled, and follows the grasping of the component, transport and release phases. The transition from the end of the melting phase to the grasping phase is challenging due to several factors, namely the presence of smaller components in the surrounding space causing the component of interest to constrain the natural motion promoted by the tool contact during desoldering. In all tests, the desoldering success rate is approximately 100\%, while for the grasping phase is about 50\%.

\section{Conclusion}
This work demonstrated that it is feasible to robotize the desoldering process and extracting components from PCBs, contributing to automate the recycling process of e-waste. The proposed tool design, aided by robot force control, can successfully desolder by contact PCB components from the board and transport them. Our system contemplates 6 distinct phases, approach, contact, melting, grasping, transport and release. Here, the force control successfully acted to keep the set contact force from the approach to the melting phase, guaranteeing that the tool is in constant contact with the component. The grasping phase showed to be more challenging to achieve due to the small size of the components and the other components in the surrounding attached to the board. Tests performed in different PCBs showed the versatility of the system and its ability to desolder the bigger components, even from smaller PCBs like the ones of mobile phones. The system can remove the PCB components with a high success rate, approximately 100\% for the bigger PCB components.

Future work will be dedicated to improving the system by adding a second robotic arm to the hot air application and the integration of a computer vision system to detect the PCB components of interest (their pose and classification).

\subsubsection*{Acknowledgements}
This work is funded by National Funds through the FCT - Fundação para a Ciência e
Tecnologia, I.P., under the scope of the project RECY-SMARTE - Sustainable approaches for
recycling and re-use of discarded mobile phones (PTDC/CTA-AMB/3489/2021).
%
%

\end{document}